\documentclass[12pt]{article}
\usepackage[margin=1in]{geometry}
\usepackage[utf8]{inputenc}
\usepackage{xcolor}
\usepackage{amsmath}
\usepackage{apacite}
\usepackage{graphicx}

\usepackage{svg}
\graphicspath{ {figures/} } 

\title{A neural network walks into a lab: towards using deep nets as models for human behavior}

\author{Wei Ji Ma and Benjamin Peters}
\date{\today}

\begin{document}

\maketitle

\begin{abstract}
What might sound like the beginning of a joke  has become an attractive prospect for many cognitive scientists: the use of deep neural network models (DNNs) as models of human behavior in perceptual and cognitive tasks. Although DNNs have taken over machine learning, attempts to use them as models of human behavior are still in the early stages. Can they become a versatile model class in the cognitive scientist's toolbox? We first argue why DNNs have the potential to be interesting models of human behavior. We then discuss how that potential can be more fully realized. On the one hand, we argue that the cycle of training, testing, and revising DNNs needs to be revisited through the lens of the cognitive scientist's goals. Specifically, we argue that methods for assessing the goodness of fit between DNN models and human behavior have to date been impoverished. On the other hand, cognitive science might have to start using more complex tasks (including richer stimulus spaces), but doing so might be beneficial for DNN-independent reasons as well. Finally, we highlight avenues where traditional cognitive process models and DNNs may show productive synergy.
\end{abstract}

\section*{Highlights}

\begin{itemize} % max 900 characters
%• Highlights are a short collection of bullet point statements (3-5) that concisely convey to the reader the recent advances in the area, including emerging concepts and/or distinctions, that constitute a main motivation for the discussion developed in the article. 
%• As Highlights focus on recent developments, conclusions and future directions should instead be discussed in the Concluding Remarks section and/or the Outstanding Questions box.

    \item Like their predecessors -- connectionist models -- DNNs should be taken seriously as candidate models of behavior. 
    \item DNN models of behavior would ideally be trained in two stages: ecological pre-training and training on diagnostic tasks. Possibly a third, consisting of limited fitting to behavioral data. 
    \item For evaluating the goodness of fit of a DNN, just like of any model, as rich as possible a characterization of behavior should be used. Reaction times and learning trajectories are underused for this purpose.
    \item The computational power of DNNs should stimulate the development of more complex human tasks, which can then also pose new challenges for cognitive process models.
    \item Hybrids of cognitive process models and DNNs offer new opportunities.
\end{itemize}
\newpage

\tableofcontents

\section{Introduction}

Deep neural networks (DNNs) currently dominate machine learning. Since their breakthrough performance in visual object recognition \cite{krizhevsky_imagenet_2012}, they are approaching or surpassing human performance across a wide range of complex real-world tasks \cite{lecun_deep_2015}. Increased computational power and the availability of large-scale data sets have made it possible to train more complex (e.g., deeper) models. In addition, the neural network toolkit has been expanded by new computational building blocks like attention \cite{vaswani2017attention}, graph neural networks \cite{battaglia2018relational}, read/write memory \cite{graves2014neural}, and the fusion with probabilistic modeling \cite{kingma_auto-encoding_2013, rezende_stochastic_2014}.

To machine learners, human performance often serves merely as a benchmark for \textbf{deep neural network} (see Glossary) performance, and making DNNs behave similarly to humans is not their goal. However, in recent years, DNNs have attracted the attention of cognitive scientists and cognitive neuroscientists, who would like to turn them into good models of human behavior and neural activity. For them, the performance of a DNN on a \textbf{task} is not interesting in itself. Having a DNN that behaves at or above human levels is only the starting point for digging deeper: can the network account quantitatively for the richness of human behavior across task conditions and task parameters?

As deep neural network models are entering the cognitive science toolkit, it is worthwhile to formulate a research strategy through which DNNs could become more valuable models of human behavior across a range of tasks, and challenges faced by this strategy. In particular, cognitive scientists should not blindly adopt the training and testing methodologies of DNNs from machine learning, since those were developed with different goals in mind.

\begin{figure}[htp]
    \centering
    \includegraphics[width=16cm]{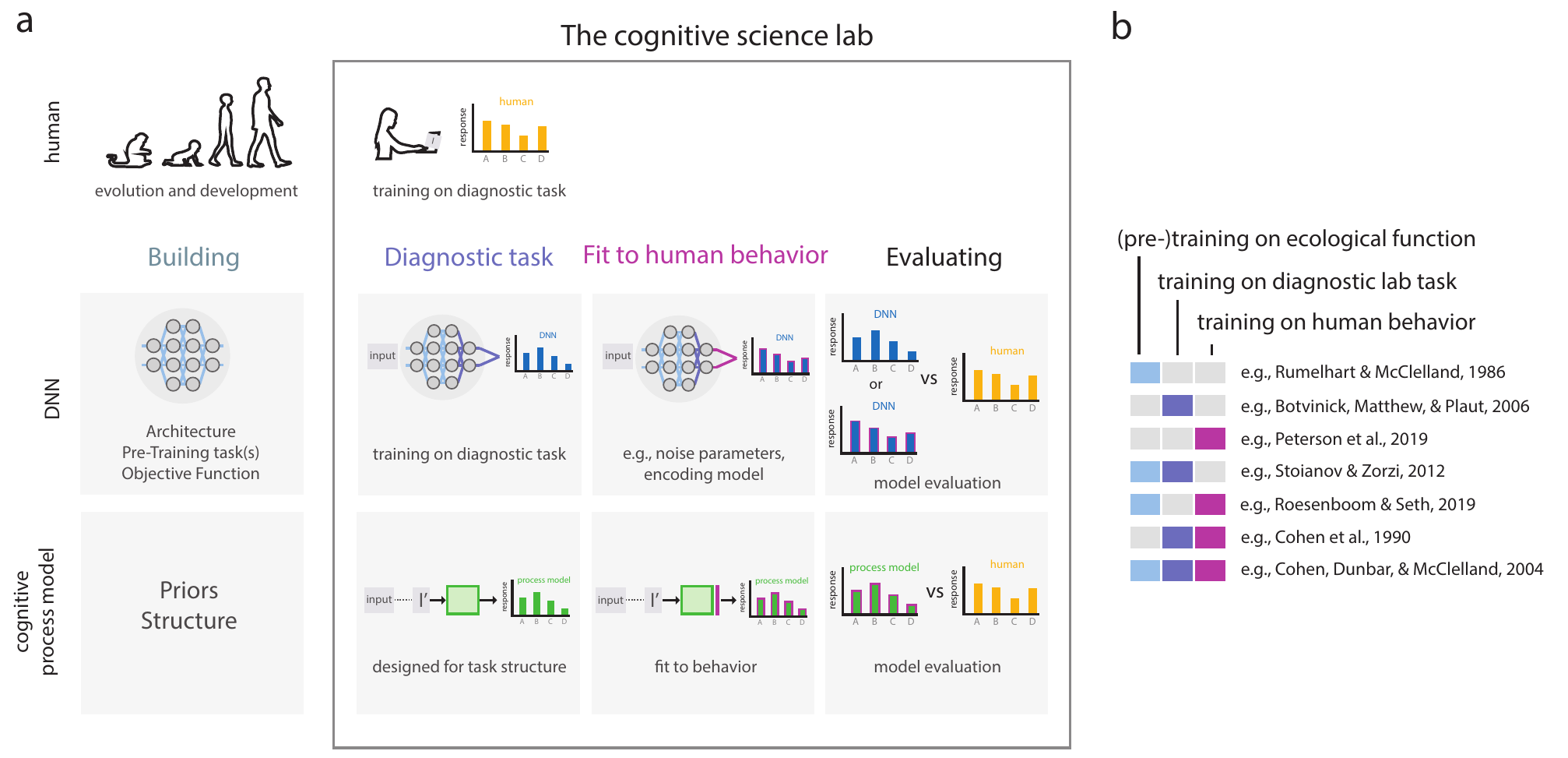}    
    \caption{A DNN walks into a lab -- a metaphor. (a) First row: The human mind is the result of evolutionary and individual development and learning. Cognitive scientists devise diagnostic tasks to test computational hypotheses of the mind. Second row: The architecture, training schedule, and the pre-training tasks are building blocks that shape the DNN model's ``mind" to become a testable hypothesis of a human cognitive function (Section \ref{sec:building}). This function is then being tested by diagnostic lab tasks (i.e., cognitive experiments) (Section \ref{sec:testing}). In the process, (parts of) the network may be trained to perform the diagnostic task (akin to a human that is trained on a lab task) and some aspects of the DNN might be fit to human behavior (e.g., encoding model parameters or noise parameters) (Section \ref{sec:increasing}). Rich behavioral measures should be used to evaluate the viability of the model (Section \ref{sec:evaluating}). Third row: In contrast to DNNs, cognitive process models more explicitly define the building blocks of the model (priors, process structure). They are usually specifically devised for the diagnostic task and take low-dimensional, hand-crafted representations ($I'$) as input. (b) The proposed research strategy involves three different training stages (pre-training, training on diagnostic lab task, fitting to human behavior). Different combinations of these have been employed in the literature.}.
    \label{fig:paperoverview}
\end{figure}

In this paper, we attempt to formulate conditions under which DNN models could become interesting and relevant models of human behavior for cognitive science and cognitive neuroscience. How should we train and test DNN models? How do DNN models compare and contrast with the cognitive process models that are the bread and butter of cognitive science? We start by addressing two obvious criticisms one could have of the broad research agenda (Section \ref{obvious}). Then, we introduce the running thread of this paper: the metaphor of a neural network that - like a human participant - walks into the cognitive science lab to be studied by a cognitive scientist (see Figure \ref{fig:paperoverview}). 

A human participant entering a lab experiment has already been shaped by factors such as neural architecture, environmental statistics, development, and evolutionary pressure. These factors impose a substrate of cognitive priors onto which new experiences are grafted. To gain insight into the black box of the human participant's mind, the cognitive scientist develops a ``diagnostic" lab task to test cognitive theories and to compare participants (e.g., patients vs. controls). The way in which the participant performs the task is a function of the task training as well as their cognitive priors. To yield DNN models that behave more human-like and can therefore serve as models of human behavior, we here advocate for an analogous view on DNN models. In particular, we want DNN models that have similar ``cognitive priors" as humans. One way to achieve this is to let the neural architecture be informed by biological knowledge and to mimic development and evolution by selecting ecologically relevant training tasks (Section \ref{sec:building}). Such a hopefully ``human-like" model then enters the cognitive science lab to be tested like a human; this may involve task-specific training on the diagnostic lab task (Section \ref{sec:testing}).

At this point, the metaphor breaks down, as the cognitive scientist now wants to compare human and DNN behavior (Section \ref{sec:evaluating}) and potentially revise the model by fitting some parts of the DNN directly to human behavior or even revise the model entirely (Section \ref{sec:increasing}). Toward the end of the paper, we review some directions to integrate cognitive process models and DNN models (Section \ref{sec:towards}).

\section{The obvious criticisms}\label{obvious}
Before we go into the technicalities of making neural networks more useful as models of human cognition, we will address two broad questions one might have about this research agenda. 

\subsection{Do we really need more models in cognitive science?}\label{cogpro}
There is a rich tradition of ``hand-crafted'' cognitive process models to describe human behavior, such as evidence accumulation models, associative learning models, and Bayesian models (reviewed in \cite{ratcliff2008diffusion, miller1995assessment, ma2019bayesian}, respectively). \textbf{Cognitive process models} are typically assemblies of simple, interpretable, and purportedly universal components, such as sensory noise, linear filters, decision criteria, and error signals. Many cognitive process models are to some extent normative, in the sense that the decision-maker minimizes a cost function. Cognitive process models typically have fewer than 20 parameters, and those parameters are usually fitted to individual subjects. 

Cognitive process models often require handcrafting \textbf{feature representations} to reduce the complexity of the input to a few low-dimensional feature dimensions. For example, one might handcraft a canonical color space and a one-dimensional circular space to represent the color and orientation features of natural image patches. In planning, one might choose a set of simple state features to construct a value function \cite{sutton2018reinforcement}. In terms of the decision stage of a cognitive process model, modelers typically restrict themselves to considering a small number of decision rules that are more or less principled. This handcrafting introduces potentially unverifiable assumptions and may prevent scaling to more complex and high-dimensional tasks. These disadvantages are less present in another class of cognitive models, \textbf{connectionist models} \cite{oreilly_computational_2000, rumelhart_parallel_1986, mcclelland_letting_2010}. Such models are neural networks typically consisting of a few tens to hundreds of artificial neurons. They have been used as models for a wide range of cognitive tasks, such as past-tense learning \cite{rumelhart1986learning}, the Stroop effect \cite{cohen_control_1990}, and serial recall \cite{botvinick_short-term_2006}. Compared to handcrafted process models, they can learn much more flexible input-output mappings from data, in particular intermediate, hidden feature representations. From a modeler's perspective, this comes with the challenge to understand the emergent representations. 
As training neural network models involves gradually changing parameters, the phenomenology that a neural network displays during training may serve as a model of development or task learning \cite{rogers_semantic_2004, rosenberg_spacing_1986, rumelhart_learning_1986, saxe_mathematical_2019, munakata_rethinking_1997, munakata_connectionist_2003, yermolayeva_connectionist_2014, lake_building_2017,orhan_efficient_2017, kruschke1992alcove}. Moreover, manipulating the statistics of the task a neural network is trained on allows one to systematically study the adaptive value of its mechanisms \cite{rahwan_machine_2019, cichy_deep_2019, kietzmann_deep_2018}.

Deep neural networks inherit many properties of connectionist models like their distributed representations or end-to-end training via backpropagation. Technological advances and larger datasets have allowed them to be trained much deeper and to more complex tasks \cite{schmidhuber2015deep}. Most current state-of-the-art DNNs are typically not designed to be models of human cognition. Yet, we argue in this paper that with some adjustments to the pipeline, they can become good models of human cognition that complement traditional cognitive process models. On the one hand, cognitive models still have advantages. They allow for principled testing of specific computational hypotheses, whereas in DNNs, representations and computation are an emergent property of \textbf{task-based training}. Like in connectionist models, their computational principles are therefore often harder to understand and interpret than those of handcrafted process models. The attempt to model human behavior with a neural network has therefore been described by the attempt to substitute one black box (the human mind) with another black box. 

On the other hand, DNNs offers more possibilities than cognitive process models. First, a researcher is not confined to their imagination: the emergent properties of DNNs sometimes allow for the discovery of solutions that one might not have thought of. Second, DNNs are routinely trained on tasks that are many orders of magnitude more complex than any task described by cognitive process models; examples include DNNs for object recognition \cite{krizhevsky_imagenet_2012} and \textbf{deep reinforcement learning} models for chess and Go \cite{silver_general_2018, silver_mastering_2017}. Thus, they have the potential to vastly expand the scope of computational modeling in cognitive science, and inspire new cognitive process models.
Third, for cognitive neuroscientists, DNN models promise the ability to ``look under the hood", i.e. to understand how neural processes may give rise to behavior. Indeed, the emergent internal representations of DNNs have been shown to be good predictors of neural representations in brains \cite{khaligh-razavi_deep_2014, yamins_performance-optimized_2014,yamins_using_2016} suggesting that insights gleaned from DNNs can potentially inform cognitive neuroscience. For these reasons, we believe that DNNs might enrich the cognitive scientist's toolbox. Instead of viewing neural networks as a competitor to cognitive process models, they can complement each other in efforts to understand human behavior. Towards the end of this paper, we will address ways in which DNN models could be combined with cognitive process models.\\

%[TODO: REFER TO TABLE. CHECK IF WE ROUGHLY ADDRESS ALL ENTRIES IN THE TABLE]

\noindent
\begin{tabular}{|p{1.2in}|p{1.5in}|p{1.5in}|p{1.5in}|}
\hline
     & {\bf cognitive process models} & {\bf connectionist models} & {\bf DNN models}\\
\hline
{\bf tasks applied to} & mostly simple lab tasks & mostly simple lab tasks & mostly complex real-world tasks\\
\hline
{\bf number of parameters}    & order 1-10 & order 10-100 & order $10^3-10^9$\\
%\hline {\bf epistemological goal} & interpretability & interpretability & power\\
\hline
{\bf optimization} &  to fit human data &  to a task objective &  to a task objective\\
\hline
{\bf neural data} & no account & potential account & potential account\\
\hline
{\bf learning trajectory} & usually no explicit account & potential account & potential account\\
\hline
\end{tabular}\\

\noindent

\subsection{Does superhuman performance render DNNs irrelevant as a model of human behavior?}

According to the universal function approximation theorem, any sufficiently deep and sufficiently large network, given sufficient training data, learns to approximate any (continuous) function from input to output arbitrarily well \cite{cybenko1989approximation, hornik1991approximation}. Indeed, in some complex tasks, such as Go \cite{silver_mastering_2017} or detecting breast cancer \cite{mckinney_international_2020}, DNN performance is superhuman. Doesn't that make the DNN irrelevant as a model of human behavior? 

We argue that it does not. First, cognitive scientists have long been comfortable with models that perform better than humans. Namely, ideal-observer models describe how an observer with full information should behave to obtain maximum performance on a particular task. In many domains however, human behavior does not follow an ideal-observer model (e.g. \cite{tjan_human_1995}). These models only start to capture human behavior when augmented with limitations such as encoding noise \cite{faisal_noise_2008}, limited memory \cite{ma_changing_2014}, or  inference approximations \cite{beck2012not}. Similar to ideal-observer models in cognitive science, neural networks can provide an upper bound on the performance that can be achieved given a task and architectural constraints. Modifying a superhuman DNN in order to achieve a better fit to human behavior could take many forms, including changing the architecture, the cost function, the learning scheme, or the training data (see also Section \ref{subsec:iterativeimprovement}). 

Second, ``superhuman" is a term that is thrown around quite easily but sometimes reflects a rather narrow notion of performance. Specifically, DNNs often perform well on tasks similar to the ones on which they were trained - often on par or better than humans. A sensible requirement of a ``superhuman" model is that it also performs well on tasks and distributions that are considered similar by humans. Those models, however, very often fail to generalize outside the distribution and task they have been trained on. In general, DNNs can easily learn to solve tasks with strategies that are very different from humans (e.g., memorization) \cite{zhang_understanding_2016}. Unlike humans - such a model will fail to generalize to new stimuli outside of the training distribution. A striking and prominent example of a lack of out-of-distribution generalization of DNNs is their vulnerability to adversarial attacks. Here, targeted, changes of the input that are imperceptible to humans can yield drastic changes in the output of a DNN \cite{szegedy_intriguing_2014}. Moreover, DNNs trained to classify images with a certain noise level fail to generalize to new, untrained noise levels \cite{geirhos_comparing_2018, jo_measuring_2017}. DNNs that play Atari games at or above human-level performance display catastrophic performance drops when game objects are shifted by a few pixel compared to the training, demonstrating that these DNNs do not have an understanding of visual objects in the game \cite{kansky_schema_2017}. 

In sum, in some narrow task regimes with large amounts of training data, DNNs are incredibly good in fitting input-output relationships to an extent that may be superhuman; then, making the DNN more human-like is still interesting and potentially challenging. However, more often, DNNs do not even perform at human levels when performance is considered broadly.

\section{Designing and training more human-like DNN models}
\label{sec:building}

Building DNN models requires the researcher to select configurations in a large space of modeling dimensions. How should the architecture of a DNN be designed, which task(s) should be used to train the network, what should be the cost function \cite{marblestone2016toward}, and what should be the training procedure? 
%A systematic exploration in this space may be guided by the quest for ultimate causal explanations of behavior \cite{cichy_deep_2019, tinbergen1963aims}, e.g. why a system has evolved to be a certain way.
We here focus on exploring this space with the long-term goal of building DNN models that behave like humans on ecologically valid tasks.

%We are considering the long-term goal of building DNN models that behave like humans on ecologically valid tasks. To achieve this goal, several fundamental questions need to be answered. How should the architecture of a DNN be designed, which task(s) should be used to train the network, what should be the cost function \cite{marblestone2016toward}, and what should be the training procedure? We here highlight some promising directions and results.

\subsection{Architecture and objectives}

On the architectural side, more biologically plausible neural networks may yield more human-like behavior. The basic functionality of DNNs is loosely inspired by biological neuronal networks and it is unclear what level of biological realism is necessary to yield models displaying human-like behavior. However, in general, moving DNNs towards more biologically realistic modes of processing has seemed to be a viable strategy so far. For instance, instead of using deep feedforward networks that might have hundreds of layers, \textbf{recurrent neural networks} with lateral and feedback connections in shallower networks yield more human brain-like representations and behavior \cite{kubilius_brain-like_2019, spoerer_recurrent_2019, kietzmann_recurrence_2019}. Feedback connections are ubiquitous in the human visual system and are thought to support generative inference \cite{clark_whatever_2013} and DNNs that perform generative inference display robustness to adversarial stimuli \cite{schott_towards_2018}. These models are particularly interesting to both cognitive scientists and machine learning researchers as they promise an account of self-supervised learning of the input from unlabeled data \cite{henaff2019data}. Other potential directions involve separating excitatory and inhibitory neurons \cite{song2016training}, spiking neural networks \cite{tavanaei2018deep}, oscillations \cite{reichert_neuronal_2013}, sparsity constraints \cite{boutin2019meaningful}, attention \cite{lindsay2018biological, kruschke1992alcove}, more biologically plausible learning rules \cite{lee2015difference, lillicrap2016random}, and more biologically plausible cost functions \cite{marblestone2016toward}.

While DNNs excel at learning input-output relationships from data, they tend to struggle in learning models from data that flexibly map input to variables. This may put models at an advantage that allow for relational structures \cite{battaglia2018relational}, relational memory \cite{santoro2018relational}, and dynamic mapping of information via attention \cite{bahdanau2014neural, luong2015effective}, transformer networks \cite{vaswani2017attention}, spatial transformers \cite{jaderberg2015spatial} or more explicitly build them in like Capsule networks \cite{sabour2017dynamic, hinton2018matrix}. While these new DNN mechanisms are highly interesting to machine learning researchers, it remains an open question whether their applications in DNNs will also entail a better fit to human behavior.

\subsection{The importance of tasks}\label{subsec:importanceoftasks}

Finding the right tasks to train networks is critical, as tasks are the experiences that shape the computations and representations in a network \cite{richards_deep_2019}. In cognitive process models, prior experiences may be captured by explicit priors to represent the statistics of the world which the brain has adapted to \cite{geisler_visual_2008}, or implicit priors imposed by the structure of the cognitive process model. 
By contrast, computations in standard DNNs are an emergent property of learning; DNNs are usually not equipped with explicit priors. In addition, implicit priors imposed by the architecture do not necessarily line up with the structure of the task. Instead, during task-based training, a neural network is learning the computational processes and the priors from experience. This means that in order to yield similar representations, computations, and priors as the human brains, special care has to be taken to expose the DNN to the ``correct" experiences (i.e., task set).

There are two issues that determine whether a particular task imposes the same kind of solution space on a function that a DNN should learn as evolution and real-world task demands have imposed on the human cognitive function that we want to model. 
The first is complexity: computational solutions may drastically differ when faced with higher complexity and many solutions may not scale up from low- to high- complexity regimes. Each task can be solved by a variety of different computational solutions, even more so if the task has low complexity and the potential solution space is large (e.g., a large DNN). Hence, a DNN trained on a overly simplified toy version of a task may learn a particular computational strategy that is unlike the human solution or does not generalize to the real-world complexity of the task. Hence, it is crucial to use tasks that approach real-world complexity or to at least show that a neural network function is able to scale up to real-world complexity. 
The second is modularity \cite{yang_how_2019}: in humans, a particular function might not only support a single task but may be optimized to support multiple tasks. To attain such a functional modularity, it might be necessary to train DNNs on multiple tasks. 
We now explore three aspects of task training that may address these challenges of complexity and modularity: richer training sets, pre-training with ecologically valid task, and multi-task training.

\subsection{Richer training sets}

The universal function approximation property of DNNs makes them powerful statistical pattern detectors. With adequate training, they are able to detect and exploit systematic relationships in the training distribution and generalize well to stimuli from the same distribution (i.e., typically the held-out test data). However, generalization beyond the training distribution often fails spectacularly, as discussed above. % (i.e., adversarial examples, geihros, pixel shifting in atari). 
Moreover, a large body of research on object classification DNNs has shown that DNNs can pick up any subtle correlation in the data to solve the tasks and that these may be very different from those used by humans to perform the same task \cite{ilyas_adversarial_2019, lapuschkin_unmasking_2019}. Such ``cheats" are discovered if those subtle correlations are specific to the training dataset and revealed in new datasets where those correlations do not hold \cite{barbu2019objectnet}. 

One approach to sidestep the problem of insufficient out-of-distribution generalization is to enrich the input distribution. Such data augmentation techniques are standard in the training of DNNs \cite{shorten_survey_2019}. For instance, in image recognition, the training dataset can be enhanced by identity-preserving transformations such as scaling, rotation, or adding noise to the stimulus. The DNN is then trained to give the same response to those augmented samples as the original sample. This approach does not solve the problem of generalizing beyond the training distribution. However, given sufficient rich training distributions (e.g., training on a wide variety of noise levels and variations of the input), this may yield models which perform well enough on new stimuli which are ``close enough'' to the training distribution to allow for generalization.

The strategy of enriching the input distribution does not just make for better learning but also might bring DNNs closer to humans in their learning experience. The human experience is characterized by large input variability: functions such as object recognition have to be robust across an enormous range of different input noise levels, lighting condition, viewpoints, rotations, occlusions -- all occurring during dynamic changes in the input due to external, self-, and saccadic motion. Hence, training DNNs on stimuli like videos (e.g., on head-mounted cameras \cite{sullivan2020saycam}), which approach the variability which humans were exposed to during evolution and development, might yield models with human-like robustness \cite{hasson2020direct}.

\subsection{Pre-training with ecologically valid tasks}

An attractive and the arguably most promising approach to build DNN models of human behavior is to train models to perform the same tasks as humans and to analyze similarities and differences in their behavior \cite{yamins_using_2016}. Whether this approach will yield success crucially depends on the task. Models tend to be trained from scratch to perform a particular task (e.g., digit classification). In contrast, human brains are the result of evolutionary and developmental optimization on ecologically valid tasks. These pre-existing representations and processes shape how they learn new tasks. For instance, when faced with a new task (e.g., an experimental task in a cognitive science lab) humans might perceive a connected group of pixel on a lab monitor as an object or an agent, and use intuitive physics and intuitive psychology inference \cite{lake_human-level_2015} to learn to perform this task. Untrained DNNs do not have any of these prior experiences and therefore learn new tasks using potentially vastly different strategies. 

Hence, to yield DNNs that behave and learn similarly to humans, it might be necessary for models to be exposed to a similar kind of ecological pre-training. The human brain is optimized for adaptive behavior in a dynamic and complex environment. Hence, requiring a similar pre-training for DNNs is a daunting task that might be fulfilled only in the long-term. However, all is not lost. Identifying ecologically valid sub-tasks that a modular structure of the human brain is performing might suffice in many situations \cite{richards_deep_2019}. For example, the most successful class of DNN models are based on image classification DNNs. The success of this approach in finding behavior similar to humans and neural representations similar to the primate ventral stream \cite{khaligh-razavi_deep_2014} likely results from the fact that object recognition is a ecologically valid task that the brain (i.e., the ventral stream) has been optimized for during the course of evolution and development. Ultimately, one research goal may be to identify ecologically valid brain tasks their success in inducing human-like behavior of DNNs trained on them.

\subsection{Multi-task training}

The human mind certainly is not dissociated into separable sub-functions that can each be purely taxed by a single task. Instead, multiple cognitive functions subserve performance in the same task and the same cognitive function is used across multiple tasks. Hence, attempting to identify a singular task to train a DNN may be a inherently limited approach. Instead, one can train a DNN on a set of multiple tasks that are thought to rely on a common set of sub-functions \cite{sinz_engineering_2019, yang_how_2019, collobert_unified_2008, zamir_taskonomy:_2018}. This multi-task training demands sharing of representations and computations across all tasks in a DNN. This, in turn, may support the emergence of modular sub-functions that yield human-like generalization and robustness \cite{yang_task_2019, caruana_learning_1998}.

A DNN can be trained on multiple tasks simultaneously by interleaving them in the training set. Alternatively, tasks can be trained sequentially. A particular case in which humans learn sequentially is when they start from smaller and simpler tasks and build up to more complex tasks. Inspired by this, curriculum learning \cite{bengio2009curriculum} attempts to structure training tasks in a way that may allow neural networks to learn faster and yield more human-like representations and behavior.
 
Training DNNs to perform multiple tasks is an active field of research in machine learning \cite{ruder_overview_2017}. Neural networks are faced with a particular challenge during sequential learning, as they tend to ``catastrophically forget" previously learned tasks when being trained on new tasks. However, recent progress in continual and transfer learning \cite{csurka2017domain, PARISI201954} has outlined promising strategies to overcome catastrophic forgetting, for example by avoiding changes to weights that are important for previously learned tasks \cite{kirkpatrick_overcoming_2017}.

\subsection{Inter-individual variability}
A defining nature of human cognition is its variable instantiation across different individuals. A rich tradition of inter-individual psychology research has described how humans vary in their abilities, preferences and characteristics of information processing. So far, little effort has been made to model inter-individual differences in human abilities using DNNs. One way to model this would be to train multiple instances of the same DNN architecture with different initialisations \cite{rajalingham_large-scale_2018, colunga_lexicon_2005, Mehrer2020}, different amounts or orders of experiences (i.e., training samples), different learning schedules, architectural variations (e.g., different number of neurons), or variations in cost functions (i.e., different weighting of exploration and exploitation).

\section{Testing DNN models like human subjects}\label{sec:testing}

Once a trained DNN enters the cognitive science lab, it is as much a black box as a human subject. The scientist's goals are to characterize the behavior of the DNN and to compare it with humans. To achieve these goals, they can use the same experimental tasks as they use to probe a particular cognitive function in human subjects \cite{ritter_cognitive_2017}. Because such tasks are generally different from the tasks used for ecological pre-training of DNNs, we will refer to them as \textit{diagnostic tasks}. Diagnostic tasks are typically more carefully designed than test tasks in machine learning. For example, stimulus selection in cognitive science tends to be less arbitrary than in machine learning, and often involves parametric variation. Cognitive scientists who try to understand DNNs would do well to import their tried-and-true experimental methods rather than to adopt the tests common in machine learning. In this section, we discuss how experimental techniques from cognitive science can be applied to DNNs, and how a DNN can be adjusted to perform the diagnostic task. In the next section, we will ask how to compare DNNs to humans when both have completed a diagnostic task.

\subsection{Stimulus selection} \label{subsec:stimulusselection}
Experimental approaches that are commonplace in cognitive science are not yet widely used for testing DNNs. This is particularly apparent in the selection of stimuli. Stimulus selection is necessary because the space of all possible inputs is often intractably large (e.g., a high-dimensional pixel space). In machine learning, one might attempt to test the DNN on a representative sample from the relevant stimulus distribution (e.g., a collection of photos). By contrast, a cognitive scientist would sample stimuli in a range where human behavior is most sensitive to variations (i.e, dynamic range) or most informative about the cognitive function of interest. This selection process can be informed  by  cognitive theories or previous experiments. Cognitive scientists have begun to apply the same approach to DNNs. For example, DNNs have been tested on stimuli that induce special percepts in humans, like illusions \cite{zeman_muller-lyer_2013, watanabe_illusory_2018} or perceptual completions \cite{tang_recurrent_2018}. Another important component of experimental design in humans is the parametric variation of stimulus variables. Such variation has in recent years also been applied to DNN training and testing. For example, in images, one can vary noise level \cite{geirhos_comparing_2018, wichmann_methods_2017}, clutter level \cite{chen_eccentricity_2017}, contrast \cite{geirhos_comparing_2018, leibo_psychlab:_2018, wichmann_methods_2017}, or motion coherence \cite{leibo_psychlab:_2018}. 
A stimulus selection approach that is inaccessible in the study of humans is to find stimuli that best distinguish the responses of different DNN models \cite{golan_controversial_2019}.
% Walker

\subsection{Adjusting a pre-trained DNN for a diagnostic task} \label{subsec:trainingondiagnostictask}

We have advocated for pre-training a DNN on tasks that the human brain is optimized to perform \cite{yamins_performance-optimized_2014, richards_deep_2019}. This however might yield DNN outputs that are hard or impossible to compare to human behavior. For example, the cost function during pre-training may involve semi-supervised training \cite{stoianov_emergence_2012} such as minimizing its own prediction error under a internal generative model of the world on ecologically realistic stimuli \cite{lotter_deep_2017, hinton1995wake}. Prediction errors cannot be measured directly in human behavior. Hence, in these cases humans and DNNs need to be compared on another task (i.e., the diagnostic task) such as perceptual discrimination or recognition. Parts of the pre-trained network may therefore need modification or retraining in order to perform the diagnostic task. This stage therefore corresponds to humans and models - both with their individual pre-training - being trained to perform the diagnostic task in the lab.

Recall that a researcher would like to make a claim about the human-like behavior of the pre-trained DNN. In particular, one would like to claim that the combination of architecture, cost function and pre-training gave rise to computations and representations that generalize to human-like behavior on the diagnostic task. However, given sufficient modification and re-training, any pre-trained DNN might very easily fit the diagnostic task perfectly - unlike the human subjects. In this case, the diagnostic task would become an uninformative measure of human-like behavior of the pre-trained network. Hence, retraining should be done with as little as possible training data from the diagnostic task, potentially considering a human-like amount of diagnostic task training. In addition, only the architectural modifications that are necessary to perform the task should be applied and trained. For example, one might train the weights of a simple linear classifier that maps from the higher-level representation of the (now fixed) pre-trained network to the response categories of the diagnostic task (e.g. \cite{stoianov_emergence_2012}). 

\section{Evaluating the similarity between DNN and humans}
\label{sec:evaluating}
By now, a DNN has participated in our diagnostic laboratory task and produced data. We then want to characterize how its behavior differs from that of humans, just like an experimental psychologist might compare groups of humans. This requires us to decide what metrics to use for comparison.

\subsection{Behavior beyond accuracy}
In machine learning, DNNs are often evaluated using a single metric, such as classification accuracy or playing strength. The approach of the cognitive scientist should be different. A DNN might achieve the same accuracy as a human observer in the same task, but with a very different pattern of errors. Therefore, we need to characterize behavior in richer detail than using overall accuracy. The more detailed the characterization, the stronger the test of the DNN as a model of human behavior. 

How can we achieve this? The relationship between the response $r$ of a human or a DNN to a stimulus $s$ can be formalized by an input-response distribution $p(r|s)$. The response $r$ could be multi-dimensional, for example, consisting of a decision, a confidence rating, and a time series of eye fixation locations. Across repetitions of the same stimulus $s$, humans typically vary in their response; this response variability is captured by $p(r|s)$. Neural networks can be made similarly stochastic, for example through a softmax read-out, using bayesian neural networks \cite{gal_dropout_2016}, or by injecting noise at an earlier stage. 

The most comprehensive metric to compare human and stochastic DNN behavior would be based on the full response distribution $p(r|s)$: DNN behavior is similar to human behavior if for each stimulus, their conditional response distributions are similar:
\begin{align}\label{kldiv}
    p_\text{human}(r|s) \approx p_\text{DNN}(r|s).
\end{align}
Formally, dissimilarity could be measured by the Kullback-Leibler (KL) divergence \cite{cover_elements_2012} between $p_\text{human}$ and $p_\text{DNN}$. KL divergence is a “distance” between probability distributions and is related to statistical deviance. Minimizing the KL divergence between data and a model is equivalent to maximizing the log likelihood of the model.

Using KL divergence as a metric may, however, not always be desirable. First, KL divergence is often difficult to estimate. Second, KL divergence would not reveal {\it how} the DNN and the human differ, only {\it by how much}. The full input-response distribution of humans and models is determined by a wide range of factors, and some of them might be subordinate to the phenomenon studied from the perspective of the researcher; for example, human response variability may result from factors such as attentional lapses or motor errors that may be ``uninteresting" to a researcher interested in object recognition. Considering only projections of the input-response distribution, i.e., summary statistics, allow the researcher to focus on the aspects of the input-response distribution which are most relevant for the research question at hand. Therefore, to answer the ``how" question, we need a suite of summary statistics of behavior.

\begin{figure}[htp]
    \centering
    \includegraphics[width=16cm]{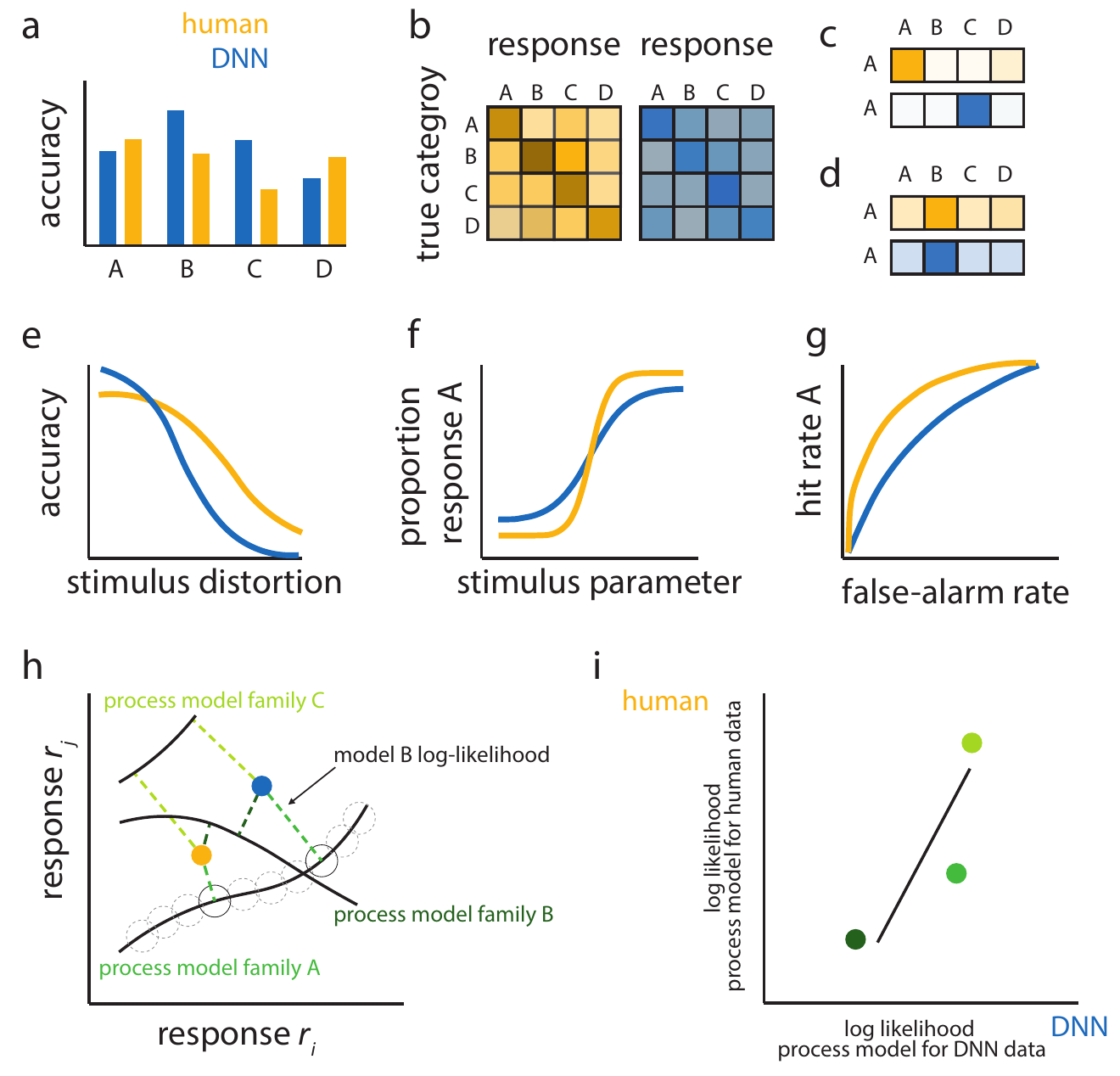}    
    \caption{The cognitive scientist's toolbox. Humans and models can be compared via accuracy (\textbf{a}) or confusion matrices across stimulus classes (\textbf{b}). Particular stimuli, such as adversarially pertubed stimuli (\textbf{c}) and illusiory stimuli (\textbf{d}) induce specific confusion patterns. Via parametric manipulation of stimuli one can compare psychometric curves (\textbf{e}, \textbf{f}). Receiver operator characteristic (\textbf{g}). Process model ranking (\textbf{h}): a process model is a family of distributions $p_\theta(\cdot|s)$ (indicated by a  black line and gray dotted circles). Moving along the black line could correspond to fitting nuisance parameters $\theta$ (e.g., a lapse parameter) to maximize the likelihood of a response vector $r$ (e.g., responses of a model and human in blue and yellow respectively). In the case of isotropic $p_\theta(\cdot|s)$, $p_\text{ML}(r|s)$ under the maximum likelihood model (indicated by solid circle) can intuitively be understood as the distance of $r$ to the model family in response space. These model distances induce a process model ranking (\textbf{i}) which can be compared for humans and models.}.
    \label{fig:toolbox}
\end{figure}

\subsection{The cognitive scientist's toolbox}
Cognitive scientists already have a toolbox of summary statistics that can be applied to DNNs (Fig. \ref{fig:toolbox}). A summary statistic is computed from $p(r|s)$ by some combination of aggregating across different $s$, aggregating across different $r$, and summarizing the distribution by a scalar. Below we list summary statistics that are potentially useful for comparing human and DNNs.

\paragraph{Accuracy and confusion matrix.} In a classification task, overall accuracy is a coarse summary statistic, as it lumps together different true classes (Fig. \ref{fig:toolbox}a). More informative is a confusion matrix, which specifies for each true class the probability that the human or the DNN responds any class \cite{kheradpisheh_humans_2016, rajalingham_large-scale_2018} (Fig. \ref{fig:toolbox}b); a high off-diagonal probability means that one class is easily mistaken for another. Adversarial attacks focus on a row in the confusion matrix where the DNN has a high off-diagonal entry but humans do not (Fig. \ref{fig:toolbox}c). In human behavioral experiments, it is possible to obtain a variant of the confusion matrix by asking subjects to judge the similarity between pairs of classes \cite{jozwik_deep_2017, peterson_adapting_2016}. Subjective similarity is then treated as a proxy for the probability of a confusion error; however, unlike confusion matrices, the resulting “similarity” matrix is necessarily symmetric. Some studies have tested whether DNNs are subject to the same illusions as humans, such as the Müller-Lyer illusion \cite{zeman_muller-lyer_2013} or illusory motion \cite{watanabe_illusory_2018}. Not unlike an adversarial attack, an illusion can be considered a systematic misperception of the stimulus, i.e. a specific off-diagonal element in the confusion matrix (Fig. \ref{fig:toolbox}d). When stimulus parameters are varied, one could plot visual classification accuracy or the confusion matrix as a function of those parameters \cite{geirhos_comparing_2018, wichmann_methods_2017, chen_eccentricity_2017, leibo_psychlab:_2018, wichmann_methods_2017}, and the slope of such a psychometric curve represents the observer’s sensitivity to the parameter \cite{carnevale_dynamic_2015, mante_context-dependent_2013, song_reward-based_2017, stoianov_emergence_2012}. 

\paragraph{Receiver operating characteristic.} Accuracy and response probabilities are not only affected by stimulus features, but also by varying the relative rewards associated with a correct response across classes or by varying the base rates of the classes \cite{greene_visual_2016}. In binary classification, these manipulations allow one to plot the hit rate against the false-alarm rate; the resulting curve is called a receiver-operating characteristic and can be compared between human and DNN \cite{lerer_learning_2016} (Fig. \ref{fig:toolbox}g). An alternative way to obtain a receiver-operating characteristic is to ask for a confidence rating in addition to a class response. However, confidence ratings can also be compared between human and DNN on a trial-by-trial basis, and this is potentially more informative.

\paragraph{\bf Reaction times.} Processing in DNNs lacks a sensible notion of time, as the ``hardware" differs between brains and computers. In contrast to feedforward models, recurrent neural networks do have a notion of sequentiality in processing and can be constructed to trade off speed and accuracy in their decision making \cite{spoerer_recurrent_2019}. To model human reaction times using DNNs, one might provide the (noise-augmented) output of a DNN as input to an evidence accumulation process \cite{ratcliff_theory_1978} to simultaneously model choices and reaction times. A subsequent scaling and shifting of the DNN reaction times can then fit this distribution to human reaction times \cite{cohen_control_1990}.

\paragraph{\bf Learning trajectory.} So far, we have discussed comparing behavior of a trained network. However, one can argue that the learning curve itself should be included in the comparison between DNN models and humans \cite{ratcliff_connectionist_1990}. Specifically, DNNs tend to need large numbers of labeled examples. This is an unrealistic model of human learning \cite{lake_building_2017}. To do better, each of the summary statistics discussed above can be measured along the learning trajectory. For example, one could require that a DNN model does not only match human accuracy after training, but also during training when presented with the same training examples as the human. In practice however, it is difficult to establish what training examples humans receive during development. 
Nevertheless, DNNs can replicate inductive biases found in humans, such as the shape bias in object recognition \cite{colunga_lexicon_2005,feinman_learning_2018, ritter_cognitive_2017}. Along similar lines, in the realm of perceptual learning, DNNs can be tested for the same qualitative properties as observed in humans, such as increased specificity \cite{wenliang_deep_2018}. Other efforts that focused on learning dynamics have moved beyond conventional DNNs to unsupervised training \cite{anselmi_unsupervised_2016,eslami_neural_2018, flesch_focused_2018, lotter_deep_2017, stoianov_emergence_2012,watanabe_illusory_2018} or to architectures with strong prior knowledge \cite{lake_human-level_2015}.

\subsection{Advanced techniques}

\paragraph{Using cognitive process models as tools.} So far, we have discussed model-free ways of characterizing behavioral data. We could go a step forward by fitting a cognitive process model to both the DNN data and human data, and comparing the parameter estimates \cite{wang2016learning}. A similar approach is common in computational studies of development or psychiatric disorders. If more than a single cognitive process model is available, we have an additional possibility: fit all such models to both sets of data, use a goodness-of-fit metric such as AIC, and compare the rankings of the process models on this metric between DNN and human \cite{orhan_efficient_2017} (Fig. 2h,i). 

\paragraph{Turing test.} Humans and DNNs can  be put in “generative” mode, in which they produce samples (such as images or speech) from a given class label – e.g. “draw a rabbit” \cite{fan_common_2017, lake_human-level_2015}. For comparing human and DNN behavior on such tasks, the challenge is that the generated samples are typically high-dimensional. Similarity could be evaluated using a Turing test \cite{turing_a.m._computing_1950}: a human observer could be presented with samples generated by a human and samples generated by a DNN, and try to tell them apart \cite{lake_human-level_2015}. Success is achieved when the judge is at chance. The Turing test bypasses the explicit formulation of a summary statistic, but relies on the judge to implicitly use whatever information is available to tell the difference. The judge in a Turing test does not need to be human; it could also be another DNN that is trained to adjudicate between human and machine generated samples. In generative adversarial training, such a judge can be pitted against a generative DNN that is trained to increase its ability to fool the judge \cite{li_turing_2016}.

\section{Increasing the similarity between DNN and humans}
\label{sec:increasing}

The outcome of the aforementioned procedure of testing and evaluating DNNs and humans most likely revealed differences in human and model behavior. (If not, the researcher has probably not explored the input space thoroughly enough.) The cognitive scientist will want to seek out and characterize the conditions in which a  DNN fails to characterize human behavior. Then, in a revision step, the researcher wants to obtain a better DNN model of human behavior (last column in Fig. 1b). This can be done in different forms, which we will discuss below.

In this process, it might be helpful to identify some ``clean", standardized, fully open data sets of human participants performing a diagnostic task. Such a data set can serve as a {\it diagnostic benchmark} for DNN models of human behavior. This notion should not be confused with that of benchmarks in machine learning. There, a benchmark task typically is a set of input stimuli (e.g., images) and corresponding normative outputs (e.g., labels of object classes) on which a DNN is trained. This DNN is then evaluated on a held-out part of the same dataset. In this case, the same task is used for both training and testing the model, and the goal is to improve DNN task performance. By contrast, a diagnostic benchmark should not serve as a full training set for a DNN model (besides minor modifications, see section \ref{subsec:trainingondiagnostictask}) but as a diagnostic test of human-like behavior. In section \ref{directly}, we discuss how a DNN could be directly fit to behavioral data; this notion is closer to the notion of a benchmark in machine learning.

\subsection{Iterative improvement of task-trained DNNs} \label{subsec:iterativeimprovement}
Once we have trained a DNN on a (diagnostic) task and compared it to human behavior on the same task, we may want to increase its similarity to human behavior, i.e. bring the DNN closer to the ultimate goal expressed in Eq. \eqref{kldiv}. There are myriad ways of doing so. 
At a coarse performance level (i.e., accuracy), a DNN may potentially differ from humans by exceeding them. In this case, one could for example decrease its number of layers, decrease the number of units per layer, limit its number of training examples, impose regularization on activity or weights, inject noise, implement working memory or attentional limitations, or entirely change its architecture. Even if DNN performance is human-like at a coarse level of task performance, any of these manipulations could bring $p_\text{DNN}(r|s)$ as a whole closer to that of humans, $p_\text{human}(r|s)$ at the level of more nuanced evaluation of model and human behavioral similarity (see section \ref{sec:evaluating}). After every change, one would retrain the network on the task and again evaluate. One could iterate this process, either within a single study or across studies, thus effectively implementing a global search in the space of all DNNs, driven by maximizing the similarity to human behavior.

\subsection{Fitting DNNs directly to behavioral data}\label{directly}

The question arises whether the process of increasing similarity to human data could be sped up by performing neural architecture search \cite{zoph2016neural} or fitting a single DNN directly on human data, instead of training it to perform a task well. Training on human behavior directly may provide a powerful tool to learn human-like priors. The hope is that DNNs would not just learn a complicated mapping between a particular stimulus and a particular human response (error). Instead, it might be easier for a neural network to develop internal priors that are similar to those human priors which gave rise to the human response (errors) in the first place. 

Just like in cognitive process modeling, one could fit the parameters of a DNN to the human behavioral data (either trial-to-trial or aggregated). Note that training on a task is equivalent to fitting to the data of an omniscient observer, but obviously human data are different. Using behavioral data only at the model evaluation stage and fitting DNNs directly to human behavior are both extreme cases of how and when human data is used to inform a model. Fitting to behavioral data could be done at different stages of model building, training and testing, corresponding to the four rows in the matrix of Fig. 1b that contain a purple box. We will now discuss examples of these four cases separately.

First, task-specific training could be followed by a stage of training on a smaller human behavioral dataset (bottom two rows of the matrix). %\cite{kummerer_understanding_2017} %cifar-datensatz, deepgaze, 
One way to increase the similarity of DNN and human behavior is to fit some parameter of the model to human behavior \cite{cohen_control_1990}. For example, DNNs trained with cross-entropy are often over-confident in their decisions \cite{guo2017calibration}. Hence, one may want to fit scaling and shift parameters of the softmax output of a DNN to match the human data \cite{golan_controversial_2019}. 

Alternatively, one could altogether replace the task-specific training by fitting human data directly, while keeping the ecological pre-training intact (fifth row). For instance, salience prediction models attempt to predict the spatial distribution of human fixations on still images \cite{bylinskii_mit_nodate, kummerer_understanding_2017}. Others have fitted additional weights to match DNN and human similarity spaces \cite{peterson_adapting_2016}, judgments \cite{battleday_modeling_2017}, and time perception \cite{roseboom_activity_2019}.

Finally, one could replace both the ecological pre-training and the task-specific training by fitting human data directly (third row) \cite{peterson_human_2019}. However, the more one replaces task training by fitting to human data, the more one has to confront the typical limitations of behavioral data sets and the resulting redundancy in DNN models of those data. In particular, the main reason that this approach is rarely used is that we would expect a powerful DNN to fit the input-response distribution (see section \ref{subsec:upperbound} for DNNs as upper bounds). Success would be neither surprising nor necessarily insightful from the cognitive scientist's perspective. Hence, if training DNNs directly on behavior one needs to evaluate the model on different data or criteria. For example, the behavioral data may become part of the pre-training and the pre-trained DNN would then be compared to human behavior on diagnostic benchmarks that were not part of the training set. Alternatively, one might be interested in additional evaluation criteria, such as the similarity of the model to neural representations or how easy it is for specific architectures or learning regimes to fit the behavioral data. Finally, large-scale behavioral experiments in which tasks are gamified may generate sufficiently large human data sets (e.g. \cite{mitroff2015can}) that fitting a DNN to those data becomes non-trivial. 

\section{Towards a closer interplay between DNN models and cognitive process models}
\label{sec:towards}

In the previous sections, we have explored how to train, evaluate, and revise DNNs to make them plausible models of human behavior. With this in mind, we would like to return to the relation with cognitive process models that we introduced in Section \ref{cogpro}.

\subsection{Towards more complex tasks}

Before we can have any conversation about how the two types of models can be compared or combined, we need to use tasks in which both types of models are interesting. We have advocated for ecologically pre-training of DNNs (see Fig. \ref{fig:paperoverview}b). If one were to follow this, then standard tasks in cognitive science (e.g., random dot motion discrimination, Treisman-style visual search, the Stroop task, or the two-step task in planning) can be used as diagnostic benchmarks in the testing phase, often without modification. However, we believe that solely using such tasks would be a missed opportunity. 

Whereas cognitive science has long aspired to explain naturalistic behavior, it has in practice been much more conservative and used extremely simple and controlled tasks (Fig. \ref{fig:taskspace}b). DNNs are able to perform much more complex tasks, approaching real-world complexity in many domains, and as we argued in Section \ref{subsec:importanceoftasks}, rely on sufficiently complex pre-training to make them appropriate models of human behavior. We argue that this should serve as an encouragement to be more ambitious and build cognitive process models of more complex tasks. For example, while visual search is well understood and modeled for controlled lab tasks with mostly simple stimuli \cite{palmer2000psychophysics, wolfe2017five}, it is less clear how those models would operate in complex real-world scenes. Object localization is a standard task in computer vision for DNNs. Both cognitive and DNN models could be challenged by matching human search behavior in more complex and dynamic visual scenes \cite{wolfe2011visual}.

In the realm of perceptual organization, excellent models exist for determining whether two line segments, separated by an occluder, belong to the same contour \cite{geisler2009contour}, but models of detecting ownership in the presence of an arbitrary number of stacked occluders \cite{pitkow2010exact} have not been tested in humans. Although occlusion stimuli that allow for cognitive modeling would likely still have to be simpler than full images as used by image segmentation DNNs, they would be a huge step up in complexity and naturalness, and they would be not be trivial for DNNs.

A similar story applies to the cognitive science of planning. Here, the most popular traditional tasks have a state space complexity of the order of $1$ to $10$ \cite{van2019tasks}, but can handle complexity as high as $10^{16}$ \cite{van2017computational}, which is a step closer to much more complex planning tasks that have been used to train DNNs \cite{silver_mastering_2017, mnih_human-level_2015, silver_general_2018, vinyals2019grandmaster}.

In conclusion, the success of DNNs on tasks that are orders of magnitude more complex than those traditionally used in cognitive science hopefully pushes the field in developing a new set of more complex tasks that are challenging for both cognitive process models as well as DNNs (Fig. \ref{fig:taskspace}b). Doing so can additionally serve as a catalyst for cognitive science to move towards the long-term goal of explaining human behavior in the complexity of the real world.

\subsection{DNNs as an upper bound for goodness of fit}\label{subsec:upperbound}
A promising practical way to use DNNs fitted to human behavior is as an approximate upper bound on the goodness of fit of cognitive process models. Such an upper bound exists because behavior is intrinsically variable, even when the input is kept constant \cite{wichmann2001psychometric}; this upper bound is analogous to the ``noise ceiling" in models of fMRI activity \cite{lage2019methods}. Specifically, the log likelihood of a model can never exceed the negative entropy of the data \cite{cover_elements_2012}. However, this entropy is notoriously difficult to estimate \cite{wichmann2001psychometric, shen2016detailed}, especially when an experiment features non-repeated stimuli or a large number of response options on each trial. When the behavioral data set is sufficiently large, a trained DNN should offer a good practical approximation to the negative entropy, thus allowing cognitive modelers to estimate the ``absolute goodness of fit" of their model.

\subsection{Directly comparing both types of models}
Beyond using DNN models as a tool for bounding goodness of fit, we can take them more seriously and allow ourselves to reject a cognitive model if it loses the competition with a DNN model. This, is, however, conceptually more complicated. The current instantiations of DNNs and cognitive process models provide different kinds of explanation of behavioral data. %with respect to the 'how' question (REF Tinberg). 
DNNs provide high-dimensional and often hard-to-interpret explanations of how the interaction of elemental operations give rise to complex behavior. By contrast, cognitive process models provide explanations at the conceptual and algorithmic level. These differences should give us pause when directly comparing models of both classes. If a DNN fits a particular human data set consistently somewhat better (in terms of log likelihood or a related metric) than a cognitive process model, this does not necessarily mean we should discard this model altogether. For example, higher interpretability of the cognitive model could overcome a small difference in goodness of fit compared to a DNN. Conversely, cognitive process models may learn faster and generalize better (i.e., more human-like) than DNNs on particular tasks as they may be equipped with sophisticated domain knowledge (e.g., an in-built physics engine for mental simulation) that a DNN has to learn from data \cite{battaglia2013simulation}. Here, the lower similarity to human behavior of a DNN may be accepted in favor of a higher potential for relating the model to neural data.

The trade-off between goodness of fit and other model qualities such as interpretability, neural plausibility, or the importance of learning representations from data are hard to pinpoint and quantify and depend on one's personal utility function for models. Therefore, we consider this a grey area, but we think it is important for researchers to state and discuss this utility function. 

\subsection{Hybrid models}

The scientific adversarial challenge between cognitive process models and DNNs may create a state in which a task can not satisfactorily be solved by neither model class. Examples could be tasks with high complexity (thereby challenging classical cognitive process models) but also relying strongly on human prior knowledge (e.g., intuitive physics or common sense reasoning, \cite{lake_human-level_2015}) that tends to be hard for DNNs to learn from data alone.  

Instead of considering DNNs and cognitive process models as contenders we can potentially combine their strengths and form ``hybrid models'' of human behavior. The differences between DNNs and cognitive process models can be leveraged to our advantage. We will now discuss such opportunities for synergy.
%With DNNs entering the cognitive science lab, they may bring their unique strengths to the table to model human behavior.
%Specifically, DNNs and process models have complementary strengths: learning representations from complex input and interpretability. Can we bring them together?

One obvious way is to learn representations with a DNN on an ecologically valid task. In a second step, the representations learned by the DNN can then be used as input to a cognitive process model (e.g. a categorization model \cite{battleday_capturing_2019} or a model of time perception \cite{roseboom_activity_2019}). %

\begin{figure}[htp]
    \centering
    \includegraphics[width=16cm]{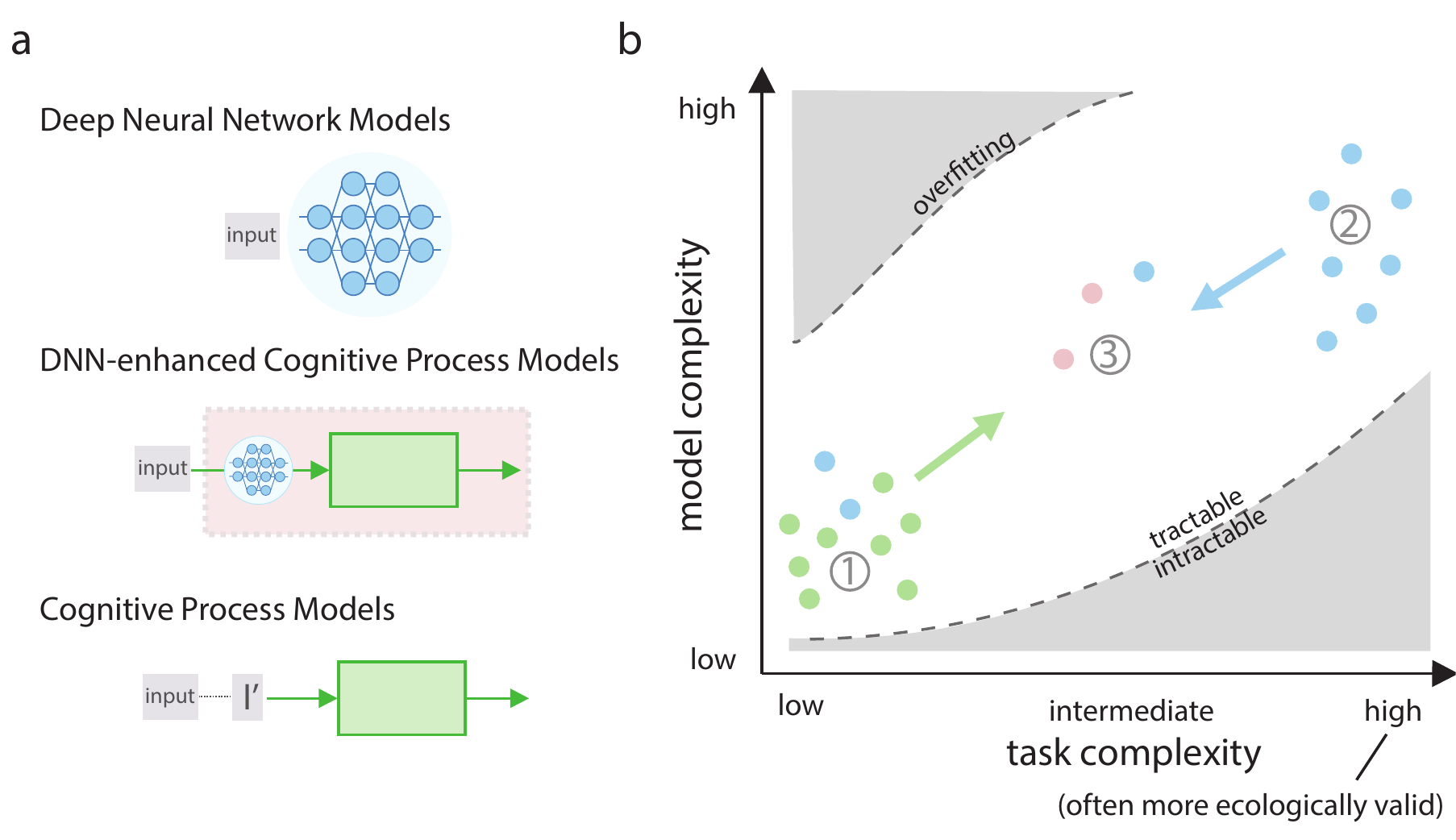}    
    \caption{Towards a closer interplay between DNNs and cognitive process models. (a) DNNs tend to be more complex than other models but are able to learn representations from data. Cognitive process models provide low-dimensional and interpretable models of human behavior. However, they require hand-crafted representations (I'). Combining both DNNs and cognitive process models may allow to learn mappings onto structured representations and therefore combine the strength of both worlds (DNN-enhanced cognitive process models). (b) A multi-faceted account of a cognitive phenomena that includes cognitive process models and DNNs presupposes that both model types are able to explain the same tasks. Complex tasks are intractable for simple models, in particular for cognitive process models (green) and low-dimensional connectionist models (blue) (1), but simple tasks may omit the potential of DNNs (blue) for real-world complexity tasks (2) and easily be overfitted by complex models. Hence, there is a large, mostly unexplored, space of tasks of intermediate complexity (3) that poses challenges for both model classes and may foster new model developments such as DNN-enhanced cognitive process models (red).}.
    \label{fig:taskspace}
\end{figure}

%If we want to bring together the two model classes, both have to be able to operate in the same task domain. Currently, this means highly controlled, low-complexity lab tasks. While our approach (pre-training on ecologically valid tasks and testing on lab-tasks) is well-suited for this, we want to challenge cognitive process models more. In particular, cognitive process models and the theories they embody need to demonstrate that they scale to real-world complex tasks. 

Cognitive process models are limited in the task complexity they can describe by the fact that representations need to be hand-crafted. However, neural networks have also been used to learn analytically intractable likelihood functions and approximate posteriors in probabilistic graphical models \cite{bengio1992global, johnson_composing_2016, kingma_auto-encoding_2013, rezende_stochastic_2014} and in a categorization task in a non-human primate \cite{walker2020neural}. Neural networks can therefore be used to learn mappings from a high-dimensional input to a low-dimensional model at the latent, cognitive level (Fig. \ref{fig:taskspace}a). In general, complementing neural networks with process-model like algorithms such as tree-search can help equip neural networks with capabilities that they would otherwise not learn from task-based training \cite{silver_general_2018} and therefore be a powerful approach for strong artificial intelligence.

There are a variety of other ways to combine DNNs and cognitive models. For instance, process models are the current standard for a model-based understanding of the behavior of the black-box which is the human mind. The same tools and can be used to better understand the black-box which is a trained DNN \cite{orhan_efficient_2017, wang2016learning} (see section \ref{sec:evaluating}). Yet, another way to use process models is to equip DNNs with inductive biases that are implemented by cognitive process models. For example, \cite{bourgin_cognitive_2019} pre-trained a DNN with a synthetic dataset generated by cognitive model and then fine-tuned the network on actual human data. 

\section{Concluding remarks}
% • Finish with clearly stated conclusions, including an indication of future directions and how any predictions you make can be tested. This section should be entitled 'Concluding remarks' or 'Concluding remarks and future perspectives'.
While connectionist models have been part of the cognitive scientist's toolbox for decades, their more powerful and popular cousin, deep neural networks, are only starting to enter the cognitive science lab. They are intriguing starting points for modeling human behavior for their ability to achieve human-level performance on complex, real-world tasks, as well as for their promise to bridge the algorithmic and implementational levels. We here outlined what a research agenda may look like to turn DNNs into good models of human behavior. We emphasized the distinction between pre-training on ecologically relevant tasks (including multi-tasking), followed by narrow training on the lab task. We then recommended using the exploration of richer tasks than common in cognitive science, as well as a rich suite of metrics to evaluate the similarity of human and DNN behavior. We discussed steps to increase this similarity, including fitting parts of the DNN directly to human data. 

We list several challenges and open questions for this research agenda in Box 2. Some of these address the scope of the data that a DNN should account for (individual differences; representation of probability; neural and behavioral data jointly), others the training of a DNN for the benefit of cognitive science (task-based training or direct fit to the data; incorporating biological detail), and yet others the types of tasks used (``toy tasks"; embodied tasks). Finally, two questions are broader model comparison questions (diagnostic power of behavior; fair comparison).

\section{Glossary}
% alphabetical order
% 'introduce the terms where relevant'
% glossary terms bold in the manuscript (when they appear the first time)
% 
\begin{itemize}

    \item \textbf{Cognitive process model}: A cognitive process model provides a mathematically specified mapping from input stimuli to subject behavior (responses, reaction times, eye movements, etc.) by combining simple, general building blocks of cognition, such as sensory noise, decision noise, priors, utility functions, and transformation rules (such as linear filters in visual processing). Each component typically comes with one or more parameters that are unknown to the experimenter. The standard way to fit a cognitive process model is to maximize the likelihood of its parameters. The likelihood of a parameter combination is the probability of generating the observed data (responses) under that parameter combination. For details, see \cite{wilson2019ten}.
 
    \item \textbf{Connectionist model}: An artificial neural network designed for the purpose of studying cognitive phenomena. Connectionist models (also known as parallel distributed processing, PDP, models) produced widespread interest in the 1980s and 1990s and have traditionally used much shallower architectures than current DNNs and employed a higher variety of learning algorithms, activation functions. The research agenda in this paper could also be termed ``new'' or ``neo-neo'' connectionism \cite{chauvet_30-year_2018} it combines the research paradigm of connectism with the computaitonal power and advances of current DNNs \cite{sep-connectionism}.

    %\item \textbf{Convolutional neural network (CNN)}: DNN in which the input has a spatial structure and which is characterized by the stacking of convolutional layers, each of which consists of feature maps that detect local correlations of features in the previous layer.     
    
    \item \textbf{Deep neural network (DNN)}: Umbrella term for an artificial neural network with a large number of hidden layers, each of which applies a sequence of simple linear and nonlinear operations to the activity in the previous layers. One example are convolutional DNNs where information is integrated spatially local with weight sharing across space. Information flows from input to the output either once (feedforward DNN) or can flow in both directions between hidden layers (recurrent DNN) where it can display complex dynamics. Recent advances like self-attention have expanded the DNN toolkit by context-dependent computations. In DNNs trained for classification, each output unit corresponds to a particular response class (e.g. target presence or object class). Current DNNs are almost exclusively trained by modyfing model weights using backpropagation of errors on a large number of training examples; DNNs are usually trained end-to-end with supervision by providing only inputs and desired outputs (e.g., correct image labels) during training, hence no manual engineering or domain knowledge is needed. The expressive power of the DNN derives from the number of transformations (i.e., layers or recurrent time-steps). 
    
    \item \textbf{Deep reinforcement learning}: A type of DNN that does not get trained using labeled examples, but using information about the positive or negative outcome of an action or action sequences – such as when winning a game. Might employ Monte Carlo tree search, a method to assess a current state (!assess the value of a potential action?) by simulating futures until an outcome is obtained.

    \item \textbf{Features}: Very often, the input is richer and higher-dimensional than needed for a task. For example, the number of pixels in an image is much higher than the number of categories that humans use for perceptual categorization. Thus, a modeler has to commit to a way of reducing the input to a set of task-relevant features. In cognitive process models, this is often done manually, based on intuition. DNNs, on the other hand, learn the features (this is also called representational learning) automatically as a part of optimizing task performance.
    
    \item \textbf{Task}: A set of stimuli combined with a prescription for how the agent should map those stimuli to a response. More generally, a mapping between a set of inputs (e.g., images) and a set of responses (e.g., labels or images).
        
    \item \textbf{Task-based training}: DNNs generally are optimized to perform a particular task. Given sufficient training data and DNN complexity, they can achieve arbitrarily good performance (universal function approximation \cite{cybenko1989approximation, hornik1991approximation}). This is different to cognitive process models which are fit to explain  human behavior.
    
    \item \textbf{Recurrent neural network (RNN)}: A DNN in which units have time-varying activation states and information can flow in cycles, e.g., bottom-up (from input to output), top-down (output to input), or lateral (within the same layer) across time. RNNs are particularly useful for time-varying sequential input but can also process static input. By 'unfolding' a RNN, it becomes equivalent to a DNN with residual connections and weight sharing across time \cite{liao2016bridging} and can be trained using the same methods as feedforward DNNs. Hence, computationally, a RNNs trades of network depth with time and often more closely corresponds to human behavior and neural data than correspondingly deep networks.
    
\end{itemize}

\newpage
\section*{Outstanding questions}
% should be phrased as questions
\begin{itemize}

    % Intro
    \item \textbf{A future for task-based training for higher cognition?} Inductive biases like convolution and hierarchical processing that have been used in computer vision (partly) coincidentally led DNNs to develop solutions similar to brain solution in object recognition. Inductive biases like attention or graph neural networks \cite{battaglia2018relational, vaswani2017attention} promise to yield success in more 'higher' cognitive tasks that require compositionality and flexible mapping of inputs to internal models of the world. Will these 'new' inductive biases similarly lead to coincidentally human-like solutions?

    % Designing
    \item \textbf{How can individual differences be captured in DNN models?} In comparing DNNs with humans, it is common to consider average human performance or to pool human data. Humans, however, have individual differences, Following cognitive process models, a DNN that purports to be a good model of human behavior should provide an account of individual differences. What will a DNN account of human inter-individual variability look like?
    
    \item \textbf{What is the role for toy tasks in neural network modeling?} In this review, we have advocated for cognitive tasks of higher complexity, primarily to pose challenges to human observers similar to the ones that machines have struggled to solve. This does not mean that we see no role for simple tasks and accompanying simple models. A toy task can represent the essence of a broad class of more complex or more real-world tasks. Understanding human or neural network behavior in a toy task can give us insights that generalize far beyond those tasks. However, we believe that ``representing the essence" and generalization should not be taken for granted. Scaling a task up in complexity (number of objects, number of interactions amongst objects, size of the state space, demands from other tasks, etc.) might force the brain or a neural network to use fundamentally different algorithms. In a given domain, what are the limitations of toy tasks and would scaling up task complexity make sense?
    
%    \item \textbf{A future for connectionism?} A parallel discussion point concerns the role of shallow %networks.
%    We have described that DNNs are in some way scaled-up, extra powerful connectionist networks. Viewed differently, however, the latter might still have advantages. Severely limiting the number of layers might help with interpretability and developing new predictions. Circumventing end-to-end learning by handcrafting task-relevant features might allow the researchers to focus on the computation of primary interest. When end-to-end learning is particularly challenging, a connectionist solution can set a benchmark for DNNs. Can connectionist networks retain a unique niche in between cognitive process models and DNNs?
    
%  \item {    While DNNs dominate machine learning, an artificial neural network does not need to be deep to account for human data in psychophysical paradigms that use strictly controlled, low-dimensional stimuli instead of natural ones (Orhan and Ma, 2017, Stoianov and Zorzi, 2012). At a mechanistic level, the resulting networks can be understood more easily than DNNs. Do such tasks and such networks have a future as toy models and tasks or do these combine the the worst of both worlds (not being powerful enough and still very hard to interpret)?
    
    \item \textbf{How sensitive are the emergent DNN models to their biological detail?} Incorporating more biological detail into DNNs will most likely increase their ability to display human-like behavior and internal activations. However, how 'smooth' is this space of biological realism - i.e., will small, critical changes in biological realism yield drastically different solutions or can we hope to adequately approximate human behavior and neural data by networks with very rudimentary biological detail? 

    \item \textbf{How may an account of probabilistic representations look like in a DNN?} Perceptual psychology has taught us that the human brain seem to represent uncertainty or even full probability distributions (reviewed in \cite{ma_neural_2014})). Bayesian probabilistic programs naturally incorporate this information \cite{lake_human-level_2015}, and probabilistic representations also emerge in trained shallow networks \cite{orhan_efficient_2017}, but does this generalize to DNNs \cite{kendall_what_2017} and can different types of uncertainty be distinguished \cite{gal_dropout_2016}?
    
    % EVALUATING
    
    \item \textbf{What are the limits of behavioral tests to discern between different models?} How far do we get with behavioral testing alone to adjudicate between competing models? This is already an important problem for cognitive process models but even more so for DNNs, as they are typically highly overparametrized. To what extent can models show the same detailed behavior but use very different algorithms and neural activations?

    % INCREASING
    
    \item \textbf{Will fitting DNNs to human behavior directly yield more human-like models?} In reference to Section \ref{directly}, can we expect more human-like out-of-distribution generalization from training DNNs directly on (large-scale) human behavior instead of the task optimal behavior? 

    \item \textbf{What are the prospects for simultaneous fitting of a DNN to behavioral and neural data}. Task-optimized DNNs are routinely “opened up” to examine the similarity of their internal representations to those of biological neurons. The same post-hoc analysis can be applied to a DNN optimized to match human data. Going further, what are the prospects for simultaneous fitting of a DNN to behavioral and neural data \cite{turner_why_2016, mcclure2016representational, kietzmann_recurrence_2019}?

    % cognitive process models vs. dnns?
    \item \textbf{What would a fair comparison between cognitive process models and DNNs look like?} When assessing the quantitative fit to human behavioral data, how should we weigh model complexity, the amount of training data, or the amount of incorporated domain knowledge when comparing cognitive process models and DNNs? In addition, the direct comparison of cognitive process models and DNN models remains rife with philosophical tension, relating to differing notions about the meaning of understanding \cite{cichy_deep_2019}. 
    
    \item \textbf{Towards embodied DNNs?} If we take the title of this paper way more seriously than anyone should, it could serve as a reminder that embodied cognition is still an under-explored area both in cognitive science and AI. Several ongoing projects try to endow virtual DNN agents with bodies that have to navigate  environments \cite{beattie2016deepmind, savva2019habitat}. Physical whole-body motor control \cite{wolpert_principles_2011} is even harder as well as the eventual integration of multiple cognitive functions such as perception and planning with motor control, such as in playing sports \cite{kitano1997robocup}.

\end{itemize}

\bibliographystyle{apacite}
\bibliography{references.bib}

\end{document}